\documentclass[11pt,a4paper]{article}
\usepackage[hyperref]{acl2017}
\usepackage{times}
\usepackage{latexsym}
\usepackage{multirow}
\usepackage{boldline}
\usepackage{amsmath}
\usepackage{graphicx}
\usepackage{subfigure}
\usepackage{url}

\aclfinalcopy 

\title{Large-scale Hierarchical Alignment for Data-driven Text Rewriting}

\author{Nikola I. Nikolov {\normalfont and} Richard H.R. Hahnloser \\
  Institute of Neuroinformatics, University of Z{\"u}rich and ETH Z{\"u}rich, Switzerland \\
  {\tt \{niniko, rich\}@ini.ethz.ch}}
  
\date{}

\begin{document}
\maketitle

\begin{abstract}

We propose a simple unsupervised method for extracting pseudo-parallel monolingual sentence pairs from comparable corpora representative of two different text styles, such as news articles and scientific papers. Our approach does not require a seed parallel corpus, but instead relies solely on hierarchical search over pre-trained embeddings of documents and sentences. We demonstrate the effectiveness of our method through automatic and extrinsic evaluation on text simplification from the normal to the Simple Wikipedia. We show that pseudo-parallel sentences extracted with our method not only supplement existing parallel data, but can even lead to competitive performance on their own.\footnote{Code available at \url{https://github.com/ninikolov/lha}.}

\end{abstract}

\section{Introduction}

Parallel corpora are indispensable resources for advancing monolingual and multilingual text rewriting tasks. Due to the scarce availability of parallel corpora, and the cost of manual creation, a number of methods have been proposed that can perform large-scale sentence alignment: automatic extraction of \textit{pseudo-parallel} sentence pairs from raw,  comparable\footnote{Corpora that contain documents on similar topics.} corpora. While pseudo-parallel data is beneficial for machine translation \cite{munteanu2005improving}, there has been little work on large-scale sentence alignment for monolingual text-to-text rewriting tasks, such as simplification \cite{nisioi2017exploring} or style transfer \cite{liustyle}. Furthermore, the majority of existing methods (e.g. \citet*{marie2017efficient,gregoire2018extracting}) assume access to some parallel training data. This impedes their application to cases where there is \textit{no parallel data available whatsoever}, which is the case for the majority of text rewriting tasks, such as style transfer. 

\begin{figure}[t!]
    \centering
    \vspace*{-3mm}
\includegraphics[width=0.48\textwidth]{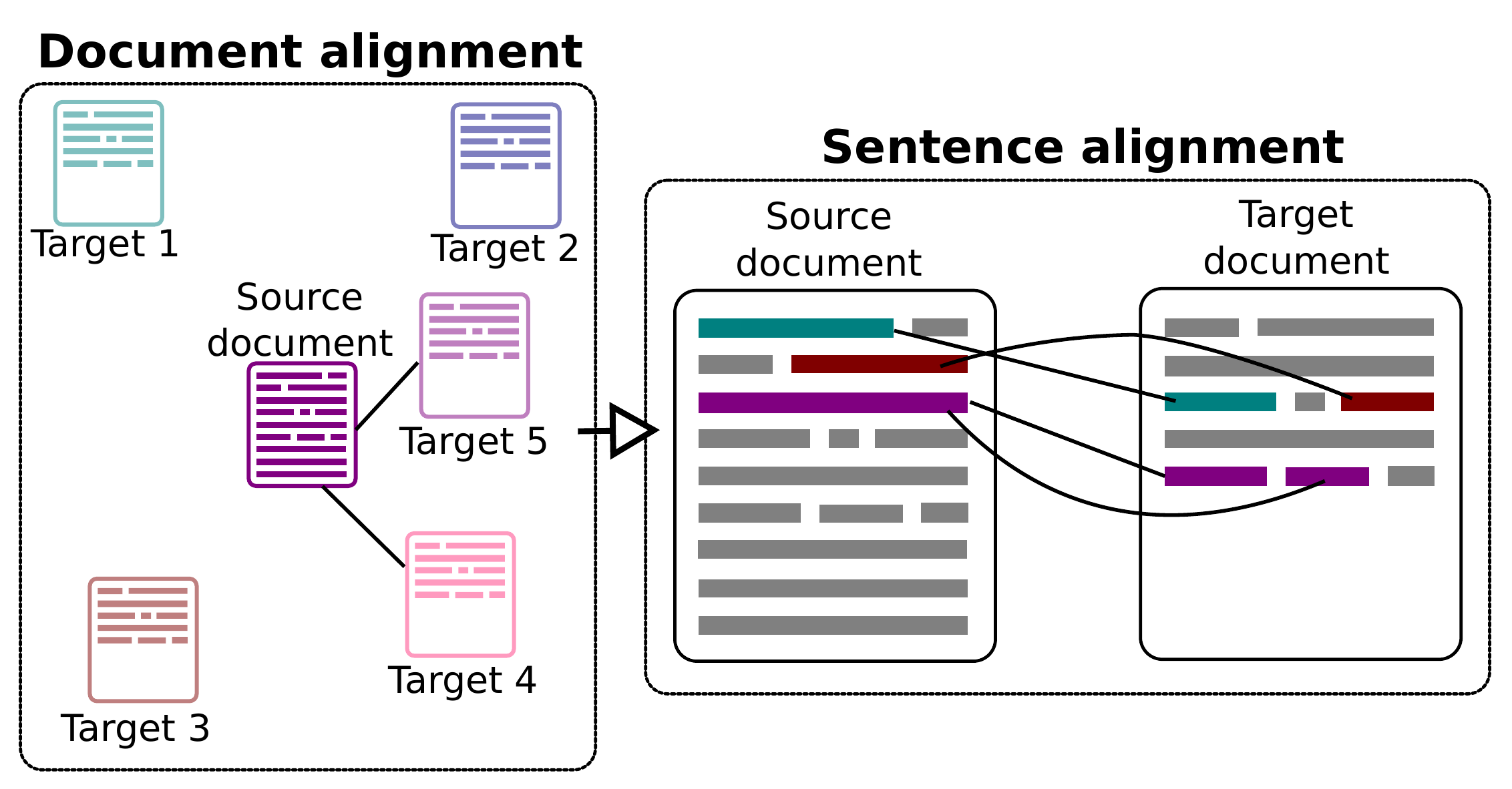}
\caption{\small{Illustration of large-scale hierarchical alignment (\textbf{LHA}). For each document in a \textit{source} dataset, \textbf{document alignment} retrieves matching documents from a \textit{target} dataset. In turn, \textbf{sentence alignment} retrieves matching sentence pairs from within each document pair.}}\label{fig:pipeline}
\vspace{-1em}
\end{figure}

In this paper, we propose a simple unsupervised method, Large-scale Hierarchical Alignment (\textbf{LHA}) (Figure \ref{fig:pipeline}; Section \ref{sec:method-large-scale}), for extracting pseudo-parallel sentence pairs from two raw monolingual corpora which contain documents in two different author styles, such as scientific papers and press releases. LHA hierarchically searches for document and sentence nearest neighbors within the two corpora, extracting sentence pairs that have high semantic similarity, yet preserve the stylistic characteristics representative of their original datasets. LHA is robust to noise, fast and memory efficient, enabling its application to datasets on the order of hundreds of millions of sentences. Its generality makes it relevant to a wide range of monolingual text rewriting tasks. 

We demonstrate the effectiveness of LHA on automatic benchmarks for alignment (Section \ref{sec:auto-evaluation}), as well as extrinsically, by training neural machine translation (NMT) systems on the task of text simplification from the normal Wikipedia to the Simple Wikipedia (Section \ref{sec:experiments}). We show that pseudo-parallel datasets obtained by LHA are not only useful for augmenting existing parallel data, boosting the performance on automatic measures, but can even be competitive on their own. 

\section{Background}\label{sec:background}

\subsection{Data-driven text rewriting}

The goal of text rewriting is to transform an input text to satisfy specific constraints, such as simplicity \cite{nisioi2017exploring} or a more general author style, such as political (e.g. democratic to republican) or gender (e.g. male to female) \cite{prabhumoye2018style,shen2017style}. Rewriting systems can be valuable when preparing a text for multiple audiences, such as simplification for language learners \cite{siddharthan2002architecture} or people with reading disabilities \cite{inui2003text}. They can also be used to improve the accessibility of technical documents, e.g. to simplify terms in clinical records for laymen \cite{abrahamsson2014medical}.

Text rewriting can be cast as a data-driven task in which transformations are learned from large collections of parallel sentences. Limited availability of high-quality parallel data is a major bottleneck for this approach. Recent work on Wikipedia and the Simple Wikipedia \cite{coster2011simple,kajiwara2016building} and on the Newsela dataset of simplified news articles for children \cite{xu2015problems} explore supervised, data-driven approaches to text simplification. Such approaches typically rely on statistical \cite{xu2016optimizing} or neural \cite{ŠTAJNER18.620} machine translation. 

Recent work on unsupervised approaches to text rewriting without parallel corpora is based on variational \cite{fu2017style} or cross-aligned \cite{shen2017style} autoencoders that learn latent representations of content separate from style. In \cite{prabhumoye2018style}, authors model style transfer as a back-translation task by translating input sentences into an intermediate language. They use the translations to train separate English decoders for each target style by combining the decoder loss with the loss of a style classifier, separately trained to distinguish between the target styles. 

\subsection{Large-scale sentence alignment}\label{sec:large-scale}

The goal of \textbf{sentence alignment} is to extract from raw corpora sentence pairs suitable as training examples for text-to-text rewriting tasks such as machine translation or text simplification. When the documents in the corpora are \textit{parallel} (labelled document pairs, such as identical articles in two languages), the task is to identify suitable sentence pairs from each document. This problem has been extensively studied both in the multilingual \citep{brown1991aligning,moore2002fast} and monolingual  \citep{hwang2015aligning,kajiwara2016building,ŠTAJNER18.630} case. The limited availability of parallel corpora led to the development of \textbf{large-scale sentence alignment} methods, which is also the focus of this work. The aim of these methods is to extract pseudo-parallel sentence pairs from raw, non-aligned corpora. For many tasks, millions of examples occur naturally within existing textual resources, amply available on the internet. 

The majority of previous work on large-scale sentence alignment is in machine translation, where adding pseudo-parallel pairs to an existing parallel dataset has been shown to boost the translation performance \cite{munteanu2005improving,uszkoreit2010large}. The work that is most closely related to ours is \cite{marie2017efficient}, where authors use pre-trained word and sentence embeddings to extract rough translation pairs in two languages. Subsequently, they filter out low-quality translations using a classifier trained on parallel translation data. More recently, \cite{gregoire2018extracting} extract pseudo-parallel translation pairs using a Recurrent Neural Network (RNN) classifier. Importantly, these methods assume that \textit{some parallel training data is already available}, which impedes their application in settings where there is no parallel data whatsoever, which is the case for many text rewriting tasks such as style transfer. 

There is little work on large-scale sentence alignment focusing specifically on monolingual tasks. In \cite{barzilay2003sentence}, authors develop a hierarchical alignment approach of first clustering paragraphs on similar topics before performing alignment on the sentence level. They argue that, for monolingual data, pre-clustering of larger textual units is more robust to noise compared to fine-grained sentence matching applied directly on the dataset level.

\section{Large-scale hierarchical alignment (LHA)}\label{sec:method-large-scale}

Given two datasets that contain comparable documents written in two different author styles: a \textbf{source} dataset $\mathbf{S^{d}}$ consisting of $N_S$ documents $\mathbf{S^{d}} = \{s^{d}_1,...,s^{d}_{N_S}\}$ (e.g. all Wikipedia articles) and a \textbf{target} dataset $\mathbf{T^{d}}$ consisting of $N_T$ documents $\mathbf{T^{d}} = \{t^{d}_1,...,t^{d}_{N_T}\}$ (e.g. all articles from the Simple Wikipedia), our approach to large-scale alignment is hierarchical, consisting of two consecutive steps: \textbf{document} alignment followed by \textbf{sentence} alignment (see Figure \ref{fig:pipeline}). 

\subsection{Document alignment}

For each source document $s^{d}_i$, document alignment retrieves $K$ nearest neighbours $\{t^{d}_{i_1},...,t^{d}_{i_K}\}$ from the target dataset. In combination, these form $K$ pseudo-parallel document pairs $\{(s^{d}_i, t^{d}_{i_1}),...,(s^{d}_i, t^{d}_{i_K})\}$. Our aim is to select document pairs with high semantic similarity, potentially containing good pseudo-parallel sentence pairs representative of the document styles of each dataset. 

To find nearest neighbours, we rely on two components: document embedding and approximate nearest neighbour search. For each dataset, we pre-compute document embeddings $e_d()$ as $I_s = [e_d(s^{d}_1),...,e_d(s^{d}_{N_S})]$ and $I_t = [e_d(t^{d}_1),...,e_d(t^{d}_{N_T})]$. We employ nearest neighbour search methods\footnote{We use the Annoy library \url{https://github.com/spotify/annoy}.} to partition the embedding space, enabling fast and efficient nearest neighbour retrieval of similar documents across $I_s$ and $I_t$. This enables us to find $K$ nearest \textit{target} document embeddings in $I_t$ for each \textit{source} embedding in $I_s$.  We additionally filter document pairs whose similarity is below a manually selected threshold $\theta_d$. In Section \ref{sec:auto-evaluation}, we evaluate a range of different document embedding approaches, as well as alternative similarity metrics. 

\subsection{Sentence alignment}

Given a pseudo-parallel document pair $(s^d, t^d)$ that contains a \textbf{source} document $s^d = \{s^s_1,...,s^s_{N_J}\}$ consisting of $N_J$ sentences and a \textbf{target} document $t^d = \{t^s_1,...,t^s_{N_M}\}$ consisting of $N_M$ sentences, sentence alignment extracts pseudo-parallel sentence pairs $(s^s_i, t^s_j)$ that are highly similar. 

To implement sentence alignment, we first embed each sentence in $s^d$ and $t^d$ and compute an inter-sentence similarity matrix $P$ among all sentence pairs in $s^d$ and $t^d$. From $P$ we extract $K$ nearest neighbours for each source and each target sentence. We denote the nearest neighbours of $s^s_i$ as $NN(s^s_i) = \{t^s_{i_1}, \dots, t^s_{i_K}\}$ and the nearest neighbours of $t^s_j$ as $NN(t^s_j)= \{s^s_{j_1}, \dots, s^s_{j_K}\}$. We remove all sentence pairs with similarity below a manually set threshold $\theta_s$. We then merge \textit{all} overlapping sets of nearest sentences in the documents to produce pseudo-parallel sentence sets (e.g. $(\{s^s_e, s^s_i\}, \{t^s_j, t^s_k, t^s_l\})$ when source sentence $i$ is closes to target sentences $j$, $k$, and $l$ and target sentence $j$ is closest to source sentences $e$ and $i$). This approach, inspired from \cite{ŠTAJNER18.630}, provides the flexibility to model multi-sentence interactions, such as sentence splitting or compression, as well as individual sentence-to-sentence reformulations. Note that when $K=1$, we only retrieve individual sentence pairs. 

The final output of sentence alignment is a list of pseudo-parallel sentence pairs with high semantic similarity and preserved stylistic characteristics of each dataset. The pseudo-parallel pairs can be used to either augment an existing parallel dataset (as in Section \ref{sec:experiments}), or independently, to solve a new author style transfer task for which there is no parallel data available (see the supplementary material for an example).

\subsection{System variants} \label{sec:system-variants}

The aforementioned framework provides the flexibility of exploring diverse variants, by exchanging document/sentence embeddings or text similarity metrics. We compare all variants in an automatic evaluation in Section \ref{sec:auto-evaluation}. 

\paragraph{Text embeddings}

We experiment with four text embedding methods: 

\begin{enumerate}
    \item \textit{Avg}, is the average of the constituent word embeddings of a text\footnote{We use the Google News 300-dim Word2Vec models.}, a simple approach that has proved to be a strong baseline for many text similarity tasks. 
    \item In \textit{Sent2Vec}\footnote{We use the public unigram Wikipedia model.} \cite{pagliardini2017unsupervised}, the word embeddings are specifically
    optimized towards additive combinations over the sentence
    using an unsupervised objective function. This approach performs well on many unsupervised and supervised text similarity tasks, often outperforming more sophisticated supervised recurrent or convolutional architectures, while remaining very fast to compute. 
    \item \textit{InferSent}\footnote{We use the GloVe-based model provided by the authors.} \cite{conneau2017supervised} is a supervised sentence embedding approach based on bidirectional LSTMs, trained on natural language inference data. 
    \item \textit{BERT}\footnote{We use the base 12-layer model provided by the authors.} \cite{devlin2018bert} is a state-of-the-art supervised sentence embedding approach based on the Transformer architecture. 
\end{enumerate}

\paragraph{Word similarity}

We additionally test four word-based approaches for computing text similarity. Those can be used either on their own, or to refine the nearest neighbour search across documents or sentences. 
\begin{enumerate}
    \item We compute the unigram string overlap $o(\mathbf{x}, \mathbf{y})$ $= \frac{|\{\mathbf{y}\} 	\cap \{\mathbf{x}\}|}{|\{\mathbf{y}\}|}$ between source tokens $\mathbf{x}$ and target tokens $\mathbf{y}$ (excluding punctuation, numbers and stopwords).
    \item We use the \textit{BM25} ranking function \cite{robertson2009probabilistic}, an extension of TF-IDF.
    \item We use the Word Mover's Distance (\textit{WMD}) \cite{kusner2015word}, which measures the distance the embedded words of one document need to travel to reach the embedded words of another document. WMD has recently achieved good results on text retrieval \cite{kusner2015word} and sentence alignment \cite{kajiwara2016building}.
    \item We use the Relaxed Word Mover's Distance (\textit{RWMD}) \cite{kusner2015word}, which is a fast approximation of the WMD. 
\end{enumerate}

\section{Automatic evaluation} \label{sec:auto-evaluation}

\begin{table*}[]
\centering
\small 
\caption{\small{Automatic evaluation of Document (left) and Sentence alignment (right). \textbf{EDim} is the embedding dimensionality. \textbf{TP} is the percentage of true positives obtained at $\mathbf{F1}_{max}$. Speed is calculated on a single CPU thread.}}
\label{tab:sent-align}
\begin{tabular}{c|c|c|c|c|c|c|c|c|c|c}
\multicolumn{3}{c}{}
& \multicolumn{4}{|c|}{\textbf{Document alignment}} & \multicolumn{4}{c|}{\textbf{Sentence alignment}} \\ \hlineB{3}
& \textbf{Approach} & \textbf{EDim} & $\mathbf{F1}_{max}$ & $\mathbf{TP}$ & $\theta_d$ & \textbf{doc/s} & $\mathbf{F1}_{max}$ & $\mathbf{TP}$ & $\theta_s$ & \textbf{sent/s} \\ \hlineB{3}
\parbox[t]{1mm}{\multirow{3}{*}{\rotatebox[origin=c]{90}{\small{\textit{Embedding}}}}} & 
Average word embeddings (Avg) & 300 & 0.66 & 43\% & 0.69 & 260 & 0.675 & 46\% & 0.82 & 1458 \\ 
 & Sent2Vec \cite{pagliardini2017unsupervised} & 600  & \textbf{0.78} & \textbf{61\%} & 0.62 & 343 & 0.692 & 48\% & 0.69 & 1710  \\ 
 & InferSent$^{\dagger}$ \cite{conneau2017supervised} & 4096 & - & - & - & - & 0.69 & 49\% & 0.88 & 110 \\ 
& BERT$^{\dagger}$ \cite{devlin2018bert} & 768 & - & - & - & - & 0.65 & 43\% & 0.89 & 25 \\ \hlineB{2}
\parbox[t]{1mm}{\multirow{4}{*}{\rotatebox[origin=c]{90}{\small{\textit{Word sim}}}}}
 & Overlap & - & 0.53 & 29\% & 0.66 & 120 & 0.63 & 40\% & 0.5 & 1600  \\ 
 & BM25 \cite{robertson2009probabilistic} & - & 0.46 & 16\% & 0.257 & 60 & 0.52 & 27\% & 0.43 & 20K  \\ 
 & RWMD \cite{kusner2015word} & 300 & 0.713 & 51\% & 0.67 & 60 & 0.704 & 50\% & 0.379 & 1050  \\ 
 & WMD \cite{kusner2015word} & 300 & 0.49 & 24\% & 0.3 & 1.5 & \textbf{0.726} & \textbf{54\%} & 0.353 & 180  \\ \hlineB{2}
 & \cite{hwang2015aligning} & - & - & - & - & - & 0.712 & - & - & -  \\ 
 & \cite{kajiwara2016building} & - & - & - & - & - & 0.724 & - & - & -  \\ \hlineB{3}
\end{tabular}
\\
    \vspace{1mm}
    {\small $\dagger$: These models are specifically designed for sentence embedding, hence we do not test them on document alignment.}
    
\vspace*{-1mm}

\end{table*}

\begin{table}[]
    \centering
    \small 
    \caption{\small{Evaluation on large-scale sentence alignment: identifying the \textit{good} sentence pairs without any document-level information. We pre-compute the embeddings and use the Annoy ANN library. For the WMD-based approaches, we re-compute the top 50 sentence nearest neighbours of Sent2Vec.}}
    \label{tab:largescale}
    \begin{tabular}{c|c|c|c}
    \hlineB{3}
    \textbf{Approach} & $\mathbf{F1}_{max}$  & $\mathbf{TP}$ & \textbf{time} \\ \hlineB{3}
    LHA (Sent2Vec) & 0.54 & 31\% & 33s \\ 
    LHA (Sent2Vec $+$ WMD) & \textbf{0.57} & \textbf{33\%} & 1m45s \\ \hline
    Global (Sent2Vec) & 0.339 & 12\% & 15s \\ 
    Global (WMD) & 0.291 & 12\% & 30m45s \\ \hlineB{3}
    \end{tabular}
    
\vspace*{-5mm}
\end{table}

We perform an automatic evaluation of LHA using an annotated sentence alignment dataset \cite{hwang2015aligning}. The dataset contains 46 article pairs from Wikipedia and the Simple Wikipedia. The 67k potential sentence pairs were manually labelled as either \textit{good} simplifications (277 pairs), \textit{good} with a \textit{partial} overlap (281 pairs), \textit{partial} (117 pairs) or \textit{non-valid}. We perform three comparisons using this dataset: evaluating document and sentence alignment separately, as well as jointly. 

For sentence alignment, the task is to retrieve the $277$ good sentence pairs out of the $67k$ possible sentence pairs in total, while minimizing the number of false positives. To evaluate document alignment, we add $1000$ randomly sampled articles from Wikipedia and the Simple Wikipedia as noise, resulting in $1046$ article pairs in total. The goal of document alignment is to identify the original $46$ document pairs out of $1046\times 1046$ possible document combinations. 

This set-up additionally enables us to \textbf{jointly} evaluate document and sentence alignment, which best resembles the target effort of retrieving good sentence pairs from noisy documents. The two aims of the joint alignment task are to identify the \textit{good} sentence pairs from within either $1M$ document or $125M$ sentence pairs, in the latter case without relying on any document-level information whatsoever. 

\subsection{Results}

Our results are summarized in Tables \ref{tab:sent-align} and \ref{tab:largescale}. For all experiments, we set $K=1$ and report the maximum F1 score ($\mathbf{F1}_{max}$) obtained from varying the document threshold $\theta_d$ and the sentence threshold $\theta_s$. We also report the percentage of true positive ($\mathbf{TP}$) document or sentence pairs that were retrieved when the F1 score was at its maximum, as well as the average speed of each approach (\textbf{doc/s} and \textbf{sent/s}). The speed becomes of a particular concern when working with large datasets consisting of millions of documents and hundreds of millions of sentences. 

On document alignment, (Table \ref{tab:sent-align}, left) the \textit{Sent2Vec} approach achieved the best score, outperforming the other embedding methods including the word-based similarity measures. On sentence alignment (Table \ref{tab:sent-align}, right), the WMD achieves the best performance, matching the result from \cite{kajiwara2016building}. When evaluating document and sentence alignment jointly (Table \ref{tab:largescale}), we compare our hierarchical approach (\textit{LHA}) to global alignment applied directly on the sentence level (\textit{Global}). \textit{Global} computes the similarities between all $125M$ sentence pairs in the entire evaluation dataset. LHA significantly outperforms Global, successfully retrieving three times more valid sentence pairs, while remaining fast to compute. This result demonstrates that document alignment is beneficial, successfully filtering some of the noise, while also reducing the overall number of sentence similarities to be computed. 

The \textit{Sent2Vec} approach to LHA achieves good performance on document and sentence alignment, while also being the fastest to compute. We therefore use it as the default approach for the following experiments on text simplification.

\section{Empirical evaluation}\label{sec:experiments}

To test the suitability of pseudo-parallel data extracted with LHA, we perform empirical experiments on text simplification from the normal Wikipedia to the Simple Wikipedia. We chose simplification because some parallel data are already available for this task, allowing us to experiment with mixing parallel and pseudo-parallel datasets. In the supplementary material 
we experiment with an additional task for which there is no parallel data: style transfer from scientific journal articles to press releases. 

We compare the performance of neural machine translation (NMT) systems trained under three different scenarios: 1) using \textbf{existing} parallel data for training; 2) using a \textbf{mixture} of parallel and pseudo-parallel data extracted with LHA; and 3) using pseudo-parallel data \textbf{on its own}. 

\subsection{Experimental setup}

\paragraph{NMT model}

For all experiments, we use a single-layer LSTM encoder-decoder model \cite{cho2015describing} with an attention mechanism \cite{bahdanau2014neural}. We train our models on the subword level \cite{sennrich2015neural}, capping the vocabulary size to 50k. We re-learn the subword rules separately for each dataset,  and train until convergence using the Adam optimizer \cite{kingma2014adam}. We use beam search with a beam of 5 to generate all final outputs. 

\paragraph{Evaluation metrics} 

\begin{table*}[]
\small
\centering
\caption{\small{Datasets used to extract pseudo-parallel monolingual sentence pairs in our experiments.}}
\label{table:datasets}
\begin{tabular}{c|c|c|c|c|c|c}
\hlineB{3}
\textbf{Dataset} & \textbf{Type} & \textbf{Documents} & \textbf{Tokens} & \textbf{Sentences} & \textbf{Tok. per sent.} & \textbf{Sent. per doc.} \\ \hlineB{3}
Wikipedia & Articles & 5.5M & 2.2B &  92M & 25 $\pm$ 16 & 17 $\pm$ 32 
\\ 
Simple Wikipedia & Articles & 134K & 62M &  2.9M & 27 $\pm$ 68 & 22 $\pm$ 34 
\\ 
Gigaword & News & 8.6M & 2.5B &  91M & 28 $\pm$ 12 & 11 $\pm$ 7 
\\ \hlineB{3}
\end{tabular}
\vspace*{-3mm}
\end{table*}

\begin{table*}[!htbp]
\centering
\caption{\small{Example pseudo-parallel pairs extracted by our Large-scale hierarchical alignment (LHA) method.}}
\label{tab:pseudo-par-examples-compact}
\small 
\begin{tabular}{p{1cm}|p{6.7cm}|p{6.7cm}}
\hlineB{3}
\multicolumn{1}{p{1cm}|}{\textbf{Dataset}} & \multicolumn{1}{c|}{\textbf{Source}} & \multicolumn{1}{c}{\textbf{Target}} \\ \hlineB{3}
\multicolumn{1}{p{1cm}|}{wiki-simp-65} & However, Jimmy Wales, Wikipedia's co-founder, denied that this was a crisis or that Wikipedia was running out of admins, saying, "The number of admins has been stable for about two years, there's really nothing going on." & But the  co-founder Wikipedia,  Jimmy Wales, did not believe  that this was a crisis. He also did not believe Wikipedia was running out of admins. \\ \hline
wiki-news-74 & Prior to World War II, Japan's industrialized economy was dominated by four major zaibatsu: Mitsubishi, Sumitomo, Yasuda and Mitsui. & Until Japan 's defeat in World War II , the economy was dominated by four conglomerates , known as `` zaibatsu '' in Japanese . These were the Mitsui , Mitsubishi , Sumitomo and Yasuda groups . \\ \hlineB{3}
\end{tabular}
\end{table*}

\begin{table}[]
\caption{\small{Statistics of the pseudo-parallel datasets extracted with LHA. $\mu_{tok}^{src}$ and $\mu_{tok}^{tgt}$ are the mean src/tgt token counts, while $\%_{s>2}^{src}$ and $\%_{s>2}^{tgt}$ report the percentage of items that contain more than one sentence.}}
\label{tab:pseudo-parallel-datasets}
\resizebox{218px}{!}{%
\begin{tabular}{c|c|c|c|c|c}
\hlineB{3}
\textbf{Dataset} & \textbf{Pairs} & \textbf{$\mu_{tok}^{src}$} & \textbf{$\mu_{tok}^{tgt}$} & \textbf{$\%_{s>2}^{src}$} & \textbf{$\%_{s>2}^{tgt}$} \\ \hlineB{3}
wiki-simp-72 & 25K & 26.72 & 22.83 & 16\% & 11\% \\ 
wiki-simp-65 & 80K & 23.37 & 15.41 & 17\% & 7\% \\ 
wiki-news-74 & 133K & 25.66 & 17.25 & 19\% & 2\% \\ 
wiki-news-70 & 216K & 26.62 & 16.29 & 19\% & 2\% \\ \hlineB{3}
\end{tabular}%
}
\vspace*{-3mm}
\end{table}

\begin{table*}
\small
\centering
\caption{\small{Empirical results on text simplification from Wikipedia to the Simple Wikipedia. The highest SARI/BLEU results from each category are in bold. \texttt{input} and \texttt{reference} are not generated using Beam Search. }}
\label{tab:wiki-simple}
\resizebox{\textwidth}{!}{%
\begin{tabular}{c|c|c|c|c|c|c|c|c|c|c|c}
\hlineB{3}
\multirow{2}{*}{\textbf{Method or Dataset}} & \multirow{2}{*}{\textbf{\begin{tabular}[c]{@{}c@{}}Total pairs \\ (\% pseudo)\end{tabular}}} & \multicolumn{5}{c|}{\textbf{Beam hypothesis 1}} & \multicolumn{5}{c}{\textbf{Beam hypothesis 2}} \\ \cline{3-12} 
 &  & \textbf{SARI} & \textbf{BLEU} & \textbf{$\mu_{tok}$} & $LD_{src}$ & $LD_{tgt}$ & \textbf{SARI} & \textbf{BLEU} & \textbf{$\mu_{tok}$} & \textbf{$LD_{src}$} & $LD_{tgt}$ \\ \hlineB{3} 
input & - & 26 & 99.37 & 22.7  & 0 & 0.26 & - & - & - & - & - \\ 
reference & - & 38.1 & 70.21 &  22.3  & 0.26 & 0 & - & - & - & - & - \\ 
NTS & 282K (0\%) & 30.54 & 84.69 & - & - & - & 35.78 & 77.57 & - & - & - \\ \hlineB{2}
\multicolumn{12}{c}{\textit{Parallel + Pseudo-parallel or Randomly sampled data (Using full parallel dataset, 282K parallel pairs)}} \\ \hlineB{2}
baseline-282K & 282K (0\%) & 30.72 & 85.71 & 18.3  & 0.18 & 0.37 & 36.16 & 82.64 &  19  & 0.19 & 0.36 \\ \hlineB{2}
 + wiki-simp-72 & 307K (8\%) & 30.2 & 87.12 & 19.43  & 0.14 & 0.34 & 36.02 & 81.13 & 19.03  & 0.19 & 0.36 \\ 
 + wiki-simp-65 & 362K (22\%) & \textbf{30.92} & \textbf{89.64} & 19.8  & 0.13 & 0.33 & 36.48 & 83.56 & 19.37  & 0.18 & 0.35 \\ 
 + wiki-news-74 & 414K (32\%) & 30.84 & 89.59 & 19.67  & 0.13 & 0.33 & 36.57 & \textbf{83.85} & 19.13  & 0.18 & 0.35 \\ 
 + wiki-news-70 & 498K(43\%) & 30.82 & 89.62 & 19.6  & 0.13 & 0.33 & 36.45 & 83.11 & 18.98  & 0.19 & 0.36 \\ \hlineB{2}
 + rand-100K & 382K (26\%) & 30.52 & 88.46 & 19.7  & 0.14 & 0.34 & \textbf{36.96} & 82.86 & 19  & 0.2 & 0.36 \\ 
 + rand-200K & 482K (41\%) & 29.47 & 80.65 & 19.3  & 0.18 & 0.36 & 34.36 & 74.67 & 18.93  & 0.23 & 0.38 \\ 
 + rand-300K & 582K (52\%) & 28.68 & 75.61 & 19.57  & 0.23 & 0.4 & 32.34 & 68.9 & 18.35  & 0.3 & 0.43 \\ \hlineB{2}
\multicolumn{12}{c}{\textit{Parallel + Pseudo-parallel data (Using partial parallel dataset, 71K parallel pairs)}} \\ \hlineB{2}
baseline-71K & 71K (0\%) & \textbf{31.16} & 69.53 & 17.45 & 0.29 & 0.44 & 32.92 & 67.29 & 19.14  & 0.3 & 0.44 \\ \hlineB{2}
 + wiki-simp-65 & 150K (52\%) & 31.0 & \textbf{81.52} & 18.26  & 0.21 & 0.38 & \textbf{35.12} & \textbf{77.38} & 18.16 & 0.25 & 0.39 \\ 
 + wiki-news-70 & 286K(75\%) & 31.01 & 80.03 & 17.82 & 0.23 & 0.4 & 34.14 & 76.44 & 17.31  & 0.28 & 0.43 \\ \hlineB{2}
\multicolumn{12}{c}{\textit{Pseudo-parallel data only}} \\ \hlineB{2}
wiki-simp-all & 104K (100\%) & 29.93 & 60.81 & 18.05  & 0.36 & 0.47 & 30.13 & 57.46 & 18.53  & 0.39 & 0.49 \\ 
wiki-news-all & 348K (100\%) & 22.06 & 28.51 & 13.68  & 0.6 & 0.63 & 23.08 & 29.62 & 14.01  & 0.6 & 0.64 \\ 
pseudo-all & 452K (100\%) & \textbf{30.24} & \textbf{71.32} & 17.82  & 0.3 & 0.43 & \textbf{31.41} & \textbf{65.65} & 17.65  & 0.33 & 0.45 \\ \hlineB{3}
\end{tabular}%
}\\
\end{table*}

We report a diverse range of automatic metrics and statistics. \textit{SARI} \cite{xu2016optimizing} is a recently proposed metric for text simplification which correlates well with simplicity in the output. SARI takes into account the total number of changes (additions, deletions) of the input when scoring model outputs. \textit{BLEU} \cite{papineni2002bleu} is a precision-based metric for machine translation commonly used for evaluation of text simplification \cite{xu2016optimizing,ŠTAJNER18.620} and of style transfer \cite{shen2017style}. Recent work has indicated that BLEU is not suitable for assessment of simplicity \cite{sulembleu}, it correlates better with meaning preservation and grammaticality, in particular when using multiple references. We also report the average Levenshtein distance (LD) from the model outputs to the input ($LD_{src}$) or the target reference ($LD_{tgt}$). On simplification tasks, LD correlates well with meaning preservation and grammaticality \cite{sulembleu}, complementing BLEU. 

\paragraph{Extracting pseudo-parallel data}

We use LHA with \textit{Sent2Vec} (see Section \ref{sec:method-large-scale}) to extract pseudo-parallel sentence pairs for text simplification. To ensure some degree of lexical similarity, we exclude pairs whose string overlap (defined in Section \ref{sec:system-variants}) is below $0.4$, and pairs in which the target sentence is more than $1.5$ times longer than the source sentence. We use $K=5$ in all of our alignment experiments, which enables extraction of up to 5 sentence nearest neighbours. 

\paragraph{Parallel data}\label{sec:wiki-par-data}

As a parallel baseline dataset, we use an existing dataset from \cite{hwang2015aligning}. The dataset consists of 282K sentence pairs obtained after aligning the parallel articles from Wikipedia and the Simple Wikipedia. This dataset allows us to compare our results to previous work on data-driven text simplification. We use two versions of the dataset in our experiments: \texttt{full} contains all 282K pairs, while \texttt{partial} contains 71K pairs, or 25\% of the full dataset.

\paragraph{Evaluation data} We evaluate our simplification models on the testing dataset from \cite{xu2016optimizing}, which consists of 358 sentence pairs from the normal Wikipedia and the Simple Wikipedia. In addition to the ground truth simplifications, each input sentence comes with 8 additional references, manually simplified by Amazon Meachanical Turkers. We compute \textit{BLEU} and \textit{SARI} on the 8 manual references. 

\paragraph{Pseudo-parallel data}\label{sec:pseudo-par}

We align two dataset pairs, obtaining pseudo-parallel sentence pairs for text simplification (statistics of the datasets we use for alignment are in Table \ref{table:datasets}). First, we align the normal \texttt{Wikipedia} to the \texttt{Simple Wikipedia} using document and sentence similarity thresholds $\theta_d=0.5$ and $\theta_s=\{0.72,0.65\}$, producing two datasets:  \texttt{wiki-simp-72} and \texttt{wiki-simp-65}. Because LHA uses no document-level information in this dataset, alignment leads to new sentence pairs, some of which may be distinct from the pairs present in the existing parallel dataset. We monitor for and exclude pairs that overlap with the testing dataset. Second, we align \texttt{Wikipedia} to the \texttt{Gigaword} news article corpus \cite{napoles2012annotated}, using $\theta_d=0.5$ and $\theta_s=\{0.74,0.7\}$, resulting in two additional pseudo-parallel datasets: \texttt{wiki-news-74} and \texttt{wiki-news-70}. With these datasets, we investigate whether pseudo-parallel data extracted from a \textit{different domain} can be beneficial for text simplification. We use slightly higher sentence alignment thresholds for the news articles because of the domain difference.

We find that the majority of the pairs extracted contain a single sentence, and $15$-$20\%$ of the source examples and $5$-$10\%$ of the target examples contain multiple sentences (see Table \ref{tab:pseudo-parallel-datasets} for additional statistics). Most multi-sentence examples contain two sentences, while $0.5$-$1\%$ contain $3$ to $5$ sentences. Two example aligned outputs are in Table \ref{tab:pseudo-par-examples-compact} (additional examples are available in the supplementary material). They suggest that our method is capable of extracting high-quality pairs that are similar in meaning, even spanning across multiple sentences. 

\paragraph{Randomly sampled pairs}

We also experiment with adding random sentence pairs to the parallel dataset (\texttt{rand-100K}, \texttt{rand-200K} and \texttt{rand-300K} datasets, containing 100K, 200K and 300K random pairs, respectively). The random pairs are uniformly sampled from the Wikipedia and the Simple Wikipedia, respectively. With the random pairs, we aim to investigate how model performance changes as we add an increasing number of sentence pairs that are non-parallel but are still representative of the two dataset styles.

\subsection{Automatic evaluation}

The simplification results in Table \ref{tab:wiki-simple} are organized in several sections according to the type of dataset used for training. We report the results of the top two beam search hypotheses produced by our models, considering that the second hypothesis often generates simpler outputs \cite{ŠTAJNER18.620}. 

In Table \ref{tab:wiki-simple}, \texttt{input} is copying the normal Wikipedia input sentences, without making any changes. \texttt{reference} reports the score of the original Simple Wikipedia references with respect to the other 8 references available for this dataset. \texttt{NTS} is the previously best reported result on text simplification using neural sequence models \cite{ŠTAJNER18.620}. \texttt{baseline-\{282K, 71K\}} are our parallel LSTM baselines, trained on 282K and 71K parallel pairs, respectively.

The models trained on a \textit{mixture} of \textit{parallel} and \textit{pseudo-parallel} data generate longer outputs on average, and their output is more similar to the \texttt{input}, as well as to the original Simple Wikipedia \texttt{reference}, in terms of the LD. Adding pseudo-parallel data frequently yields BLEU improvements on both Beam hypotheses: over the \textit{NTS} system, as well as over our baselines trained solely on parallel data. The BLEU gains are larger when using the smaller parallel dataset, consisting of 71K sentence pairs. In terms of SARI, the scores remain either similar or slightly better than the baselines, indicating that simplicity in the output is preserved. The second Beam hypothesis yields higher SARI scores than the first one, in agreement with \cite{ŠTAJNER18.620}. Interestingly, adding out-of-domain pseudo-parallel news data (\texttt{wiki-news-*} datasets) results in an increase in BLEU despite the potential change in style of the target sequence.

Larger pseudo-parallel datasets can lead to bigger improvements, however noisy data can result in a decrease in performance, motivating careful data selection. In our \textit{parallel and random} set-up, we find that an increasing number of random pairs added to the parallel data progressively degrades model performance. However, those models still manage to perform surprisingly well, even when over half of the pairs in the dataset are random. Thus, neural machine translation can successfully learn target transformations despite substantial data corruption, demonstrating robustness to noisy or non-parallel data for certain tasks. 

When training solely on \textit{pseudo-parallel} data, we observe lower performance on average in comparison to parallel models. However, the results are encouraging, demonstrating the potential of our approach in tasks for which there is no parallel data available. As expected, the out-of-domain news data (\texttt{wiki-news-all}) is less suitable for simplification than the in-domain data (\texttt{wiki-simp-all}), because of the change in output style of the former. Results are best when mixing all pseudo-parallel pairs into a single dataset (\texttt{pseudo-all}). Having access to a small amount of \textit{in-domain} pseudo-parallel data, in addition to \textit{out-of-domain} pairs, seems to be beneficial to the success of our approach.

\subsection{Human evaluation}

Due to the challenges of automatic evaluation of text simplification systems \cite{sulembleu}, we also perform a human evaluation. We asked 8 fluent English speakers to rate the grammaticality, meaning preservation, and simplicity of model outputs produced for 100 randomly selected sentences from our test set. We exclude any model outputs which leave the input unchanged. Grammaticality and meaning preservation are rated on a Likert scale from $1$ (Very bad) to $5$ (Very good). Simplicity of the output sentences, in comparison to the input, is rated following \cite{ŠTAJNER18.630}, between: $-2$ (much more difficult), $-1$ (somewhat more difficult), $0$ (equally difficult), $1$ (somewhat simpler) and $2$ (much simpler). 

The results are reported in Table \ref{tab:human-eval}, where we compare our parallel baseline (\texttt{baseline-272K} in Table \ref{tab:wiki-simple}) to our best model trained on a mixture of parallel and pseudo-parallel data (\texttt{wiki-simp-65}) and our best model trained on pseudo-parallel data only (\texttt{pseudo-all}). We also evaluate the original Simple Wikipedia references (\texttt{reference}) for comparison. In terms of simplicity, our pseudo-parallel systems are closer to the result of \texttt{reference} than is \texttt{baseline-272K}, indicating that they better match the target sentence style. \texttt{baseline-272K} and \texttt{wiki-simp-65} perform similarly to the references in terms of grammaticality, with \texttt{baseline-272K} having a small edge. In terms of meaning preservation, both do worse than the references, with \texttt{wiki-simp-65} having a small edge. \texttt{pseudo-all} performs worse on both grammaticality and meaning preservation, but is on par with the simplicity result of \texttt{wiki-simp-65}. 

In Table \ref{tab:simp-examples-compact}, we also show example outputs of our best models (additional examples are available in the supplementary material). The models trained on parallel plus additional pseudo-parallel data produced outputs that preserve the meaning of 'Jeddah' as a city better than our parallel baseline, while correctly simplifying \textit{principal} to \textit{main}. The model trained solely on pseudo-parallel data produces a similar output, apart from wrongly replacing \textit{jeddah} with \textit{islam}. 

\begin{table}[]
\centering
\small
\caption{\small{Human evaluation of the Grammaticality (\textbf{G}), Meaning preservation (\textbf{M}) and Simplicity (\textbf{S}) of model outputs (on the first Beam hypothesis). }}
\label{tab:human-eval}
\begin{tabular}{c|c|c|c}
\hlineB{3}
\textbf{Method} & \textbf{G} & \textbf{M} & \textbf{S} \\ \hlineB{3}
reference & 4.53 & 4.34  & 0.69 \\ \hlineB{2}
baseline-272K & 4.51 & 3.68 & 0.9 \\ \hline
 + wiki-simp-65 & 4.39 & 3.76 & 0.74 \\ \hlineB{2}
pseudo-all & 4.02 & 2.96 & 0.77 \\ \hlineB{3}
\end{tabular}
\vspace*{-3mm}
\end{table}

\begin{table}[]
\centering
\small 
\caption{\small{Example model outputs (first Beam hypothesis). }}
\label{tab:simp-examples-compact}
\begin{tabular}{p{1.cm}|p{5.9cm}}
\hlineB{3}
\textbf{Method} & \multicolumn{1}{c}{\textbf{Example}} \\ \hlineB{3}
input & jeddah is the \textbf{principal} gateway to mecca , islam ' s holiest city , which able-bodied muslims are required to visit at least once in their lifetime . \\ \hline
reference & jeddah is the \textbf{main} gateway to mecca , the holiest city of islam , where able-bodied muslims must go to at least once in a lifetime . \\ \hlineB{3}
baseline-282K & it is the \underline{highest} gateway to mecca , islam . \\ \hline
 + wiki-sim-65 & jeddah is the \textbf{main} gateway to mecca , islam 's holiest city . \\ \hline
 + wiki-news-74 & it is the \textbf{main} gateway to mecca , islam ' s holiest city . \\ \hlineB{3}
pseudo-all & islam is the \textbf{main} gateway to mecca , islam 's holiest city . \\ \hlineB{3}
\end{tabular}
\end{table}

\section{Conclusion}

We developed a hierarchical method for extracting pseudo-parallel sentence pairs from two monolingual comparable corpora composed of different text styles. We evaluated the performance of our method on automatic alignment benchmarks and extrinsically on automatic text simplification. We find improvements arising from adding pseudo-parallel sentence pairs to existing parallel datasets, as well as promising results when using the pseudo-parallel data on its own. 

Our results demonstrate that careful engineering of pseudo-parallel datasets can be a successful approach for improving existing monolingual text-to-text rewriting tasks, as well as for tackling novel tasks. The pseudo-parallel data could also be a useful resource for dataset inspection and analysis. Future work could focus on improvements of our system, such as refined approaches to sentence pairing. 

\section*{Acknowledgments}

We acknowledge support from the Swiss National Science Foundation (grant 31003A\_156976).

\bibliographystyle{acl_natbib}

\clearpage

\appendix
\section{Style transfer from papers to press releases}\label{sec:style-transfer}

We additionally test our large-scale hierarhical alignment (LHA) method on the task of style transfer from scientific journal articles to press releases. This task is novel: there is currently no parallel data available for it. The task is well-suited for our system, as there are many open access scientific repositories with millions of articles freely available. 

\paragraph{Evaluation dataset}

We download press releases from the EurekAlert\footnote{\url{www.eurekalert.org}} aggregator and identify the digital object identifyer (DOI) in the text of each press release using regular expressions. Given the DOIs, we query the full text of a paper using the Elsevier ScienceDirect API\footnote{\url{https://dev.elsevier.com/sd_apis.html}}. Using this approach, we were able to compose 26k parallel pairs of papers and their press releases. We then applied our sentence aligner with \textit{Sent2Vec} to extract 11k parallel sentence pairs, which we use in the evaluation below. 

\paragraph{Pseudo-parallel data} 

We used LHA with \textit{Sent2Vec} and nearest neighbour limit $K=5$ to extract pseudo-parallel sentence pairs. We align $350k$ additional press releases from \texttt{EurekAlert}, for which we were unable to retrieve the full text of a paper, to two large repositories of scientific papers: \texttt{PubMed}\footnote{\small{\url{www.ncbi.nlm.nih.gov/pmc/tools/openftlist}}}, which contains the full text of $1.5M$ open access papers, and \texttt{Medline}\footnote{\url{www.nlm.nih.gov/bsd/medline.html}}, which contains over $17M$ scientific abstracts (see Table \ref{table:datasets} for an overview of these datasets). After alignment using a document similarity threshold $\theta_d=0.6$ and a sentence similarity threshold $\theta_s=0.74$, we extracted $80k$ pseudo-parallel pairs in total (\texttt{paper-press-74} dataset). We additionally aligned \texttt{PubMed} and \texttt{Medline} to all \texttt{Wikipedia} articles using $\theta_d=0.6$ and $\theta_s = 0.78$, obtaining out-of-domain pairs for this task (\texttt{paper-wiki-78} dataset). Table \ref{tab:pseudo-parallel-datasets} contains some statistics of these pseudo-parallel datasets. 

\begin{table}[!htb]
\centering
\caption{\small{Statistics of the pseudo-parallel datasets extracted with LHA for the style transfer task. $\mu_{tok}^{src}$ and $\mu_{tok}^{tgt}$ are the mean src/tgt token counts, while $\%_{s>2}^{src}$ and $\%_{s>2}^{tgt}$ report the percentage of items that contain more than one sentence.}}
\label{tab:pseudo-parallel-datasets}
\resizebox{218px}{!}{%
\begin{tabular}{c|c|c|c|c|c}
\hlineB{3}
\textbf{Dataset} & \textbf{Pairs} & \textbf{$\mu_{tok}^{src}$} & \textbf{$\mu_{tok}^{tgt}$} & \textbf{$\%_{s>2}^{src}$} & \textbf{$\%_{s>2}^{tgt}$} \\ \hlineB{3}
paper-press-74 & 80K & 28.01 & 16.06 & 22\% & 1\% \\ 
paper-wiki-78 & 286K & 25.84 & 24.69 & 20\% & 11\% \\ \hlineB{3}
\end{tabular}%
}
\end{table}
\begin{table*}[]
\small
\centering
\caption{\small{Datasets used to extract pseudo-parallel monolingual sentence pairs in our style transfer experiments.}}
\label{table:datasets}
\begin{tabular}{c|c|c|c|c|c|c}
\hlineB{3}
\textbf{Dataset} & \textbf{Type} & \textbf{Documents} & \textbf{Tokens} & \textbf{Sentences} & \textbf{Tok. per sent.} & \textbf{Sent. per doc.} \\ \hlineB{3}
PubMed & Scientific papers & 1.5M & 5.5B & 230M & 24 $\pm$ 14 & 180 $\pm$ 98 \\ 
MEDLINE & Scientific abstracts & 16.8M & 2.6B & 118M & 26 $\pm$ 13 & 7 $\pm$ 4 \\ 
EurekAlert & Press releases & 358K & 206M & 8M & 29 $\pm$ 15 & 23 $\pm$ 13 \\ \hlineB{3}
\end{tabular}
\end{table*}
\begin{table*}[]
\small
\caption{\small{Empirical results on style transfer from scientific papers to press releases. \textbf{Classifier} denotes the percentage of model outputs that were classified to belong to the press release class. $\mu_{tok}$ is the mean token length of the model outputs, while $LD$ is the Levenshtein distance of a model output, to the input ($LD_{src}$) or the expected output ($LD_{tgt}$) sentences.}}
\label{table:paper-press}
\centering
\begin{tabular}{c|c|c|c|c|c|c|c}
\hlineB{3}
\textbf{Method} & \textbf{Total pairs} & \textbf{SARI} & \textbf{BLEU} & \textbf{Classifier} & \textbf{$\mu_{tok}$} & \textbf{$LD_{src}$} & \textbf{$LD_{tgt}$} \\ \hlineB{3}
input & - & 19.95 & 45.88 & 43.71\% & 31.54 & 0 & 0.39 \\ 
reference & - & - & - & 78.72\% & 26.0 & 0.39 & 0 \\ \hlineB{2}
\multicolumn{8}{c}{\textit{Pseudo-parallel data only}} \\ \hlineB{2}
paper-press-74 & 80K & 32.83 & 24.98 & \textbf{82.01\%} & 19.43  & 0.49 & 0.56 \\ 
paper-wiki-78 & 286K & 33.67 & 28.87 & 69.1\% & 27.06 & 0.43 & 0.53 \\ 
combined & 366K & \textbf{35.27} & \textbf{34.98} & 70.18\% & 22.98 & 0.37 & 0.49 \\ \hlineB{3}
\end{tabular}
\end{table*}

\subsection{Automatic evaluation}

Our results on this task are summarized in Table \ref{table:paper-press}, where \texttt{input} is the score obtained from copying the input sentences, and \texttt{reference} is the score of the original press release references. In addition to the previously described automatic measures, we also report the classification prediction of a Convolutional Neural Network (CNN) sentence classifier, trained to distinguish between the two target styles in this task: papers vs. press releases. To obtain training data for the classifier, we randomly sample $1$ million sentences each from the \texttt{PubMed} and \texttt{EurekAlert} datasets. The CNN model achieves $94\%$ classification accuracy on a held-out set. With the classifier, we aim to investigate to what extent the overall style of the press releases is captured by the model outputs. 

All of our models outperform the \texttt{input} in terms of SARI and produce outputs that are closer to the target style, according to the classifier. The in-domain \texttt{paper-press} dataset performs worse than the out-of-domain \texttt{paper-wiki} dataset, most likely because of the larger size of the latter. \texttt{paper-wiki} generates longer outputs on average than \texttt{reference}, which may be a source of lower classification score. The best performance is achieved by the \texttt{combined} dataset, which also produces outputs that are the closest to \texttt{input} and \texttt{reference}, in terms of the Levenshtein distance (LD). 

\section{Additional examples of aligned sentences and model outputs}

Table \ref{tab:pseudo-par-examples} contains additional example pseudo-parallel sentences, that were extracted with LHA, for both the text simplification and the style transfer task. Table \ref{tab:simp-examples} contains additional model outputs for text simplification, while Table \ref{tab:press-examples} contains model outputs for style transfer from papers to press releases. 

\onecolumn

\begin{table}
\centering
\caption{\small{Example pseudo-parallel pairs extracted by LHA.}}
\label{tab:pseudo-par-examples}
\small 
\begin{tabular}{l|p{6cm}|p{6cm}}
\hlineB{3}
\multicolumn{1}{c|}{\textbf{Dataset}} & \multicolumn{1}{c|}{\textbf{Source}} & \multicolumn{1}{c}{\textbf{Target}} \\ \hlineB{3}
wiki-simp-65 & The dish is often served shredded with a dressing of oil, soy sauce, vinegar and sugar, or as a salad with vegetables. & They are often eaten in a kind of salad, with soy sauce or vinegar. \\ \hline
wiki-simp-65 & The standard does not define bit rates for transmission, except that it says it is intended for bit rates lower than 20,000 bits per second. & These speeds are raw bit rates (in Million bits per second). \\ \hline
\multicolumn{1}{c|}{wiki-simp-65} & In Etruscan mythology, Laran was the god of war. However, among his attributes is his responsibility to maintain peace. Laran's consort was Turan, goddess of love and fertility, who was equated with the Latin Venus. Laran was the Etruscan equivalent of the Greek Ares and the Roman Mars. & '''Laran''' is the god of war and bloodlust in Etruscan mythology. He was the lover of Turan, goddess of love, beauty and fertility; Laran himself was the Etruscan equivalent of the Greek god Ares (Mars in Roman mythology). \\ \hline
\multicolumn{1}{c|}{wiki-simp-65} & However, Jimmy Wales, Wikipedia's co-founder, denied that this was a crisis or that Wikipedia was running out of admins, saying, "The number of admins has been stable for about two years, there's really nothing going on." & But the  co-founder Wikipedia,  Jimmy Wales, did not believe  that this was a crisis. He also did not believe Wikipedia was running out of admins. \\ \hline
wiki-news-74 & Prior to World War II, Japan's industrialized economy was dominated by four major zaibatsu: Mitsubishi, Sumitomo, Yasuda and Mitsui. & Until Japan 's defeat in World War II , the economy was dominated by four conglomerates , known as `` zaibatsu '' in Japanese . These were the Mitsui , Mitsubishi , Sumitomo and Yasuda groups . \\ \hline
wiki-news-74 & Thailand's coup leaders Thursday banned political parties from holding meetings or from conducting any other activities, according to a statement read on national television. "Political gatherings of more than five people have already been banned, but political activities can resume when normalcy is restored," the statement said. & `` In order to maintain law and order , meetings of political parties and conducting of other political activities are banned , '' the council said in its televised statement . `` Political activities can resume when normalcy is restored , '' it said . \\ \hline
paper-press-74 & Such studies have suggested that some pterosaurs may have fed like modern-day skimmers, a rarified group of shorebirds, belonging to the genera Rynchops, that fly along the surface of still bodies of water scooping up small fish and crustaceans with their submerged lower jaw. & Previous studies have suggested that some pterosaurs may have fed like modern-day 'skimmers', a rare group of shorebirds, belonging to the Rynchops group. These sea-birds fly along the surface of lakes and estuaries scooping up small fish and crustaceans with their submerged lower jaw. \\ \hline
paper-press-74 & Obsessions are defined as recurrent, persistent, and intrusive thoughts, images, or urges that cause marked anxiety, and compulsions are defined as repetitive behaviors or mental acts that the patient feels compelled to perform to reduce the obsession-related anxiety {[}26{]}. & Obsessions are recurrent and persistent thoughts, impulses, or images that are unwanted and cause marked anxiety or distress. \\ \hline
paper-wiki-78 & Then height and weight were used to calculate body mass index (kg/m2) as a general measure of weight status. & Body mass index is a mathematical combination of height and weight that is an indicator of nutritional status. \\ \hline
paper-wiki-78 & Men were far more likely to be overweight (21.7\%, 21.3-22.1\%) or obese (22.4\%, 22.0-22.9\%) than women (9.5\% and 9.9\%, respectively), while women were more likely to be underweight (21.3\%, 20.9-21.7\%) than men (5.9\%, 5.6-6.1\%). & A 2008 report stated that 28.6\% of men and 20.6\% of women in Japan were considered to be obese. Men were more likely to be overweight (67.7\%) and obese (25.5\%) than women (30.9\% and 23.4\% respectively). \\ \hlineB{3}
\end{tabular}
\end{table}

\begin{table}[!htbp]
\centering
\small 
\caption{\small{Examples for text simplification from Wikipedia to the Simple Wikipedia. \textit{*-H1} and \textit{*-H2} denote the first and second Beam outputs.}}
\label{tab:simp-examples}
\begin{tabular}{c|p{12cm}}
\hlineB{3}
\textbf{Method} & \multicolumn{1}{c}{\textbf{Example}} \\ \hlineB{3}
input & jeddah is the principal gateway to mecca , islam \&apos; s holiest city , which able-bodied muslims are required to visit at least once in their lifetime . \\ \hline
reference & jeddah is the main gateway to mecca , the holiest city of islam , where able-bodied muslims must go to at least once in a lifetime . \\ \hline
our-baseline-full-H1 & it is the highest gateway to mecca , islam . \\ \hline
our-baseline-full-H2 & it is the main gateway to mecca . \\ \hline
wiki-sim-65-H1 & jeddah is the main gateway to mecca , islam 's holiest city . \\ \hline
wiki-simp-65-H2 & it is the main gateway to mecca , islam 's holiest city . \\ \hline
wiki-news-74-H1 & it is the main gateway to mecca , islam \&apos; s holiest city . \\ \hline
wiki-news-74-H2 & the main gateway to mecca is islam \&apos; s holiest city . \\ \hline
pseudo-all-H1 & islam is the main gateway to mecca , islam 's holiest city . \\ \hline
pseudo-all-H2 & islam is the holiest city in the world . \\ \hline \hline 
input & this closely resembles the unix philosophy of having multiple programs each doing one thing well and working together over universal interfaces . \\ \hline
reference & this is very similar to the unix philosophy of having multiple programs doing one thing well and working together over universal interfaces . \\ \hline
our-baseline-full-H1 & this closely resembles the unix philosophy of having multiple programs each doing one thing well and working together over universal interfaces . \\ \hline
our-baseline-full-H2 & this closely resembles the unix method of having multiple programs each doing one thing well and working together over universal interfaces . \\ \hline
wiki-simp-65-H1 & this closely resembles the unix philosophy of having multiple programs each doing one thing well and working together over universal interfaces . \\ \hline
wiki-simp-65-H2 & this closely resembles the unix philosophy of having multiple programs each doing one thing well . \\ \hline
wiki-news-74-H1 & this closely resembles the unix philosophy of having multiple programs each doing one thing well and working together over universal interfaces . \\ \hline
wiki-news-74-H2 & this closely resembles the unix philosophy of having multiple programs each doing one thing well . \\ \hline
pseudo-all-H1 & this is like the unix philosophy of having multiple programs each doing one thing well and working together over universal interfaces . \\ \hline
pseudo-all-H2 & this is like the unix philosophy of having multiple programs each doing one thing well . \\ \hlineB{3}
\end{tabular}
\end{table}

\begin{table*}[!htbp]
\centering
\small 
\caption{\small{Examples for style transfer from papers to press releases.}}
\label{tab:press-examples}
\begin{tabular}{c|p{12cm}}
\hlineB{3}
\textbf{Method} & \multicolumn{1}{c}{\textbf{Example}} \\ \hlineB{3}
input & here we report evidence for this hypothesis and show that age-related declines in rppc grey matter volume better account for age-related changes in risk preferences than does age per se . \\ \hline
reference & " this means that gray matter volume accounts for age-related changes in risk attitude more than age itself " . \\ \hline
paper-press-74 & the authors suggest that age-related declines in rfc grey matter volume better account for age-related changes in risk preferences than does age per se . \\ \hline
paper-wiki-78 & " we report evidence for this hypothesis and show that age-related declines in rypgrey matter volume better account for age-related changes in risk preferences than does age per se " . \\ \hline
combined & " we report evidence that age-related declines in the brain volume better account for age-related changes in risk preferences than does age per se " . \\ \hline \hline
input & it is one of the most common brain disorders worldwide , affecting approximately 15 \% 20 \% of the adult population in developed countries ( global burden of disease study 2013 collaborators , 2015 ) . \\ \hline
reference & migraine is one of the most common brain disorders worldwide , affecting approximately 15-20 \% of the adults in developed countries . \\ \hline
paper-press-74 & it is one of the most common brain disorders in developed countries . \\ \hline
paper-wiki-78 & it is one of the most common neurological disorders in the united states , affecting approximately 10 \% of the adult population . \\ \hline
combined & it is one of the most common brain disorders worldwide , affecting approximately 15 \% of the adult population in developed countries . \\ \hlineB{3}
\end{tabular}
\end{table*}

\twocolumn 

\end{document}